\documentclass{article}
\usepackage[T1]{fontenc}
\usepackage[utf8]{inputenc}
\usepackage{float}
\usepackage{amsmath}
\usepackage{amsthm}
\usepackage{graphicx}
\usepackage{xargs}[2008/03/08]

\makeatletter
\theoremstyle{plain}
\newtheorem{thm}{\protect\theoremname}
  \theoremstyle{plain}
  \newtheorem{lem}[thm]{\protect\lemmaname}

\usepackage[final, nonatbib]{nips_2017}
\usepackage{caption}

\usepackage{url}% simple URL typesetting
\usepackage{amsfonts}% blackboard math symbols
\usepackage{nicefrac}% compact symbols for 1/2, etc.
\usepackage{microtype}% microtypography
\usepackage{algolyx}
\usepackage{flushend}

\title{Geometric Enclosing Networks}

\author{
    Trung Le \\
  Deakin University \\
  \texttt{trung.l@deakin.edu.au} \\
  \AND
  Hung Vu \\
  Deakin University \\
  \texttt{hungv@deakin.edu.au} \\
  \And
  Tu Dinh Nguyen \\
  Deakin University \\
  \texttt{tu.nguyen@deakin.edu.au} \\
  \And
  Dinh Phung \\
  Deakin University \\
  \texttt{dinh.phung@deakin.edu.au} \\
}

\@ifundefined{showcaptionsetup}{}{%
 \PassOptionsToPackage{caption=false}{subfig}}
\usepackage{subfig}
\makeatother

  \providecommand{\lemmaname}{Lemma}
\providecommand{\theoremname}{Theorem}

\begin{document}
% \nipsfinalcopy is no longer used

\maketitle\global\long\def\model{\text{GEN}}
\newcommand{\sidenote}[1]{\marginpar{\small \emph{\color{Medium}#1}}}

\global\long\def\se{\hat{\text{se}}}

\global\long\def\interior{\text{int}}

\global\long\def\boundary{\text{bd}}

\global\long\def\ML{\textsf{ML}}

\global\long\def\GML{\mathsf{GML}}

\global\long\def\HMM{\mathsf{HMM}}

\global\long\def\support{\text{supp}}

\global\long\def\new{\text{*}}

\global\long\def\stir{\text{Stirl}}

\global\long\def\mA{\mathcal{A}}

\global\long\def\mB{\mathcal{B}}

\global\long\def\mF{\mathcal{F}}

\global\long\def\mK{\mathcal{K}}

\global\long\def\mH{\mathcal{H}}

\global\long\def\mX{\mathcal{X}}

\global\long\def\mZ{\mathcal{Z}}

\global\long\def\mS{\mathcal{S}}

\global\long\def\Ical{\mathcal{I}}

\global\long\def\mT{\mathcal{T}}

\global\long\def\Pcal{\mathcal{P}}

\global\long\def\dist{d}

\global\long\def\HX{\entro\left(X\right)}
 \global\long\def\entropyX{\HX}

\global\long\def\HY{\entro\left(Y\right)}
 \global\long\def\entropyY{\HY}

\global\long\def\HXY{\entro\left(X,Y\right)}
 \global\long\def\entropyXY{\HXY}

\global\long\def\mutualXY{\mutual\left(X;Y\right)}
 \global\long\def\mutinfoXY{\mutualXY}

\global\long\def\given{\mid}

\global\long\def\gv{\given}

\global\long\def\goto{\rightarrow}

\global\long\def\asgoto{\stackrel{a.s.}{\longrightarrow}}

\global\long\def\pgoto{\stackrel{p}{\longrightarrow}}

\global\long\def\dgoto{\stackrel{d}{\longrightarrow}}

\global\long\def\lik{\mathcal{L}}

\global\long\def\logll{\mathit{l}}

\global\long\def\vectorize#1{\boldsymbol{#1}}

\global\long\def\vt#1{\mathbf{#1}}

\global\long\def\gvt#1{\boldsymbol{#1}}

\global\long\def\idp{\ \bot\negthickspace\negthickspace\bot\ }
 \global\long\def\cdp{\idp}

\global\long\def\das{\triangleq}

\global\long\def\id{\mathbb{I}}

\global\long\def\idarg#1#2{\id\left\{  #1,#2\right\}  }

\global\long\def\iid{\stackrel{\text{iid}}{\sim}}

\global\long\def\bzero{\vt 0}

\global\long\def\bone{\mathbf{1}}

\global\long\def\boldm{\boldsymbol{m}}

\global\long\def\be{\boldsymbol{e}}

\global\long\def\bff{\vt f}

\global\long\def\ba{\boldsymbol{a}}

\global\long\def\bb{\boldsymbol{b}}

\global\long\def\bc{\boldsymbol{c}}

\global\long\def\bB{\boldsymbol{B}}

\global\long\def\bx{\boldsymbol{x}}

\global\long\def\bl{\boldsymbol{l}}

\global\long\def\bu{\boldsymbol{u}}

\global\long\def\bo{\boldsymbol{o}}

\global\long\def\bh{\boldsymbol{h}}

\global\long\def\bs{\boldsymbol{s}}

\global\long\def\bz{\boldsymbol{z}}

\global\long\def\xnew{y}

\global\long\def\bxnew{\boldsymbol{y}}

\global\long\def\bX{\boldsymbol{X}}

\global\long\def\tbx{\tilde{\bx}}

\global\long\def\by{\boldsymbol{y}}

\global\long\def\bY{\boldsymbol{Y}}

\global\long\def\bZ{\boldsymbol{Z}}

\global\long\def\bU{\boldsymbol{U}}

\global\long\def\bv{\boldsymbol{v}}

\global\long\def\bn{\boldsymbol{n}}

\global\long\def\bV{\boldsymbol{V}}

\global\long\def\bI{\boldsymbol{I}}

\global\long\def\bw{\vt w}

\global\long\def\balpha{\gvt{\alpha}}

\global\long\def\bbeta{\gvt{\beta}}

\global\long\def\bmu{\gvt{\mu}}

\global\long\def\btheta{\boldsymbol{\theta}}

\global\long\def\bsigma{\boldsymbol{\sigma}}

\global\long\def\blambda{\boldsymbol{\lambda}}

\global\long\def\bgamma{\boldsymbol{\gamma}}

\global\long\def\bpsi{\boldsymbol{\psi}}

\global\long\def\bphi{\boldsymbol{\phi}}

\global\long\def\bPhi{\boldsymbol{\Phi}}

\global\long\def\bpi{\boldsymbol{\pi}}

\global\long\def\bomega{\boldsymbol{\omega}}

\global\long\def\bepsilon{\boldsymbol{\epsilon}}

\global\long\def\btau{\boldsymbol{\tau}}

\global\long\def\realset{\mathbb{R}}

\global\long\def\realn{\realset^{n}}

\global\long\def\integerset{\mathbb{Z}}

\global\long\def\natset{\integerset}

\global\long\def\integer{\integerset}

\global\long\def\natn{\natset^{n}}

\global\long\def\rational{\mathbb{Q}}

\global\long\def\rationaln{\rational^{n}}

\global\long\def\complexset{\mathbb{C}}

\global\long\def\comp{\complexset}

\global\long\def\compl#1{#1^{\text{c}}}

\global\long\def\and{\cap}

\global\long\def\compn{\comp^{n}}

\global\long\def\comb#1#2{\left({#1\atop #2}\right) }

\global\long\def\nchoosek#1#2{\left({#1\atop #2}\right)}

\global\long\def\param{\vt w}

\global\long\def\Param{\Theta}

\global\long\def\meanparam{\gvt{\mu}}

\global\long\def\Meanparam{\mathcal{M}}

\global\long\def\meanmap{\mathbf{m}}

\global\long\def\logpart{A}

\global\long\def\simplex{\Delta}

\global\long\def\simplexn{\simplex^{n}}

\global\long\def\dirproc{\text{DP}}

\global\long\def\ggproc{\text{GG}}

\global\long\def\DP{\text{DP}}

\global\long\def\ndp{\text{nDP}}

\global\long\def\hdp{\text{HDP}}

\global\long\def\gempdf{\text{GEM}}

\global\long\def\Gumbel{\text{Gumbel}}

\global\long\def\Uniform{\text{Uniform}}

\global\long\def\Mult{\text{Mult}}

\global\long\def\rfs{\text{RFS}}

\global\long\def\bernrfs{\text{BernoulliRFS}}

\global\long\def\poissrfs{\text{PoissonRFS}}

\global\long\def\grad{\gradient}
 \global\long\def\gradient{\nabla}

\global\long\def\partdev#1#2{\partialdev{#1}{#2}}
 \global\long\def\partialdev#1#2{\frac{\partial#1}{\partial#2}}

\global\long\def\partddev#1#2{\partialdevdev{#1}{#2}}
 \global\long\def\partialdevdev#1#2{\frac{\partial^{2}#1}{\partial#2\partial#2^{\top}}}

\global\long\def\closure{\text{cl}}

\global\long\def\cpr#1#2{\Pr\left(#1\ |\ #2\right)}

\global\long\def\var{\text{Var}}

\global\long\def\Var#1{\text{Var}\left[#1\right]}

\global\long\def\cov{\text{Cov}}

\global\long\def\Cov#1{\cov\left[ #1 \right]}

\global\long\def\COV#1#2{\underset{#2}{\cov}\left[ #1 \right]}

\global\long\def\corr{\text{Corr}}

\global\long\def\sst{\text{T}}

\global\long\def\SST{\sst}

\global\long\def\ess{\mathbb{E}}

\global\long\def\Ess#1{\ess\left[#1\right]}

\newcommandx\ESS[2][usedefault, addprefix=\global, 1=]{\underset{#2}{\ess}\left[#1\right]}

\global\long\def\fisher{\mathcal{F}}

\global\long\def\bfield{\mathcal{B}}
 \global\long\def\borel{\mathcal{B}}

\global\long\def\bernpdf{\text{Bernoulli}}

\global\long\def\betapdf{\text{Beta}}

\global\long\def\dirpdf{\text{Dir}}

\global\long\def\gammapdf{\text{Gamma}}

\global\long\def\gaussden#1#2{\text{Normal}\left(#1, #2 \right) }

\global\long\def\gauss{\mathbf{N}}

\global\long\def\gausspdf#1#2#3{\text{Normal}\left( #1 \lcabra{#2, #3}\right) }

\global\long\def\multpdf{\text{Mult}}

\global\long\def\poiss{\text{Pois}}

\global\long\def\poissonpdf{\text{Poisson}}

\global\long\def\pgpdf{\text{PG}}

\global\long\def\wshpdf{\text{Wish}}

\global\long\def\iwshpdf{\text{InvWish}}

\global\long\def\nwpdf{\text{NW}}

\global\long\def\niwpdf{\text{NIW}}

\global\long\def\studentpdf{\text{Student}}

\global\long\def\unipdf{\text{Uni}}

\global\long\def\transp#1{\transpose{#1}}
 \global\long\def\transpose#1{#1^{\mathsf{T}}}

\global\long\def\mgt{\succ}

\global\long\def\mge{\succeq}

\global\long\def\idenmat{\mathbf{I}}

\global\long\def\trace{\mathrm{tr}}

\global\long\def\argmax#1{\underset{_{#1}}{\text{argmax}} }

\global\long\def\argmin#1{\underset{_{#1}}{\text{argmin}\ } }

\global\long\def\diag{\text{diag}}

\global\long\def\norm{}

\global\long\def\spn{\text{span}}

\global\long\def\vtspace{\mathcal{V}}

\global\long\def\field{\mathcal{F}}
 \global\long\def\ffield{\mathcal{F}}

\global\long\def\inner#1#2{\left\langle #1,#2\right\rangle }
 \global\long\def\iprod#1#2{\inner{#1}{#2}}

\global\long\def\dprod#1#2{#1 \cdot#2}

\global\long\def\norm#1{\left\Vert #1\right\Vert }

\global\long\def\entro{\mathbb{H}}

\global\long\def\entropy{\mathbb{H}}

\global\long\def\Entro#1{\entro\left[#1\right]}

\global\long\def\Entropy#1{\Entro{#1}}

\global\long\def\mutinfo{\mathbb{I}}

\global\long\def\relH{\mathit{D}}

\global\long\def\reldiv#1#2{\relH\left(#1||#2\right)}

\global\long\def\KL{KL}

\global\long\def\KLdiv#1#2{\KL\left(#1\parallel#2\right)}
 \global\long\def\KLdivergence#1#2{\KL\left(#1\ \parallel\ #2\right)}

\global\long\def\crossH{\mathcal{C}}
 \global\long\def\crossentropy{\mathcal{C}}

\global\long\def\crossHxy#1#2{\crossentropy\left(#1\parallel#2\right)}

\global\long\def\breg{\text{BD}}

\global\long\def\lcabra#1{\left|#1\right.}

\global\long\def\lbra#1{\lcabra{#1}}

\global\long\def\rcabra#1{\left.#1\right|}

\global\long\def\rbra#1{\rcabra{#1}}

\begin{abstract}
Training model to generate data has increasingly attracted research
attention and become important in modern world applications. We propose
in this paper a new geometry-based optimization approach to address
this problem. Orthogonal to current state-of-the-art density-based
approaches, most notably VAE and GAN, we present a fresh new idea
that borrows the principle of minimal enclosing ball to train a generator
$G\left(\bz\right)$ in such a way that both training and generated
data, after being mapped to the feature space, are enclosed in the
same sphere. We develop theory to guarantee that the mapping is bijective
so that its inverse from feature space to data space results in expressive
nonlinear contours to describe the data manifold, hence ensuring data
generated are also lying on the data manifold learned from training
data. Our model enjoys a nice geometric interpretation, hence termed
\emph{Geometric Enclosing Networks} (GEN), and possesses some key
advantages over its rivals, namely simple and easy-to-control optimization
formulation, avoidance of mode collapsing and efficiently learn data
manifold representation in a completely unsupervised manner. We conducted
extensive experiments on synthesis and real-world datasets to illustrate
the behaviors, strength and weakness of our proposed GEN, in particular
its ability to handle multi-modal data and quality of generated data.
\end{abstract}

\section{Introduction}

Density estimation has been being a dominant approach in statistical
machine learning since its inception due to its sound theoretical
underpinnings and the ability to \emph{explain} the data. This translates
into the estimation of a distribution $\tilde{P}\left(\bx\right)$
as `close' as possible to the true, but \emph{unknown}, data distribution
$P_{\text{data}}\left(\bx\right)$. Among other advantages, an important
consequence of this approach is the ability to \emph{generate} data
by sampling from $\tilde{P}\left(\bx\right)$. In recent years, this
data generation need has been growing rapidly with the scale and magnitude
of modern applications. However, to this end, the computational aspect
of several existing density estimation approaches becomes problematic
(i.e., they cannot scale satisfactorily to ImageNet, a large-scale
visual dataset) \cite{goodfellow_nips17_gan_tutorial}. 

What becomes a `daring' idea is the recent success of an approach
that aims to generate data directly \emph{without} estimating $\tilde{P}\left(\bx\right)$
in analytical forms, pioneered most notably by Variational Autoencoder
(VAE) \cite{kingma_welling_iclr14_vae} and Generative Adversarial
Nets (GANs) \cite{goodfellow_etal_nips14_gan}. There are some key
differences between VAE and GAN, but in the end, both can be used
to generate data by first sampling $\bz$ i.i.d from a noise space,
then feeding $\bz$ through a \emph{generator} $G\left(\bz\right)$\footnote{The decoder $p\left(\bx\gv\bz\right)$ in case of VAE.}
parameterized by a neural net (NN). Let $P_{g}$ be the distribution
over the values of $G\left(\bz\right)$; then although $P_{g}$ is
not directly modeled, VAE and GAN's objective functions minimize a
suitable distance between $P_{g}$ and the true $P_{\text{data}}$.
Both have enjoyed enormous recent popularity due to its scalability
and most importantly, it is extremely efficient to generate data using
the generator $G$ in a single shot. Nonetheless, training VAE and
GAN is still a challenging problem and taming the model to generate
meaningful outputs is still like a black art. In particular, GAN suffers
from two issues: convergence and mode collapsing\footnote{i.e., the generator might generate all data to a single mode of $P_{\text{data}}$
while still guarantees that they have high likelihood.} \cite{goodfellow_nips17_gan_tutorial}. There has been several recent
attempts to address these problems. For example, modifying the criteria
to measure the distance between $P_{g}$ and $P_{\text{data}}$ yields
different variations of GANs and there has been a surging interest
along this line of research (e.g., fGAN \cite{nowozin_etal_nips16_fgan},
InfoGAN \cite{chen_etal_nips16_infogan}, Wasserstein GAN \cite{arjovsky_etal_arxiv17_wasserstein_gan});
others have tried to address the convergence and mode collapsing via
modifying the optimization process (e.g., UnrolledGAN \cite{metz2016unrolled}).

Kernel methods with their mature principle \cite{Vapnik:1995,Cortes:1995:SN:218919.218929}
have been broadly applied to a wide range of applications in machine
learning and data analytics. The key principle of kernel methods is
that a simple geometric shape (e.g., hyperplane or hypersphere) in
feature space when being mapped back to input space forms a set of
non-linear contours characterizing data. This principle was further
exploited in support vector clustering \cite{benhur_etal_jmlr01_svc},
wherein learning a minimal enclosing ball in the feature space can
induce a set of nonlinear contours that capture data manifolds and
clusters. There have been some recent attempts \cite{li2015generative,dziugaite2015training}
to leverage kernel methods with generative model. In nature, these
works \cite{li2015generative,dziugaite2015training} base on kernel
methods to define maximum mean discrepancy (MMD), which is a frequentist
estimator to measure the mean square difference of the statistics
of the two sets of sample{\small{}s}, and then minimizing this MMD
to diminish the divergence between $P_{g}$ and $P_{\text{data}}$.

This paper takes a radical departure from the density estimation view
to data generation problem. The goal remains the same: we aim to
train a generator $G\left(\bz\right)$ to be used to generate data
efficiently through i.i.d $\bz$ lying in any arbitrary uniform or
noise space. However, unlike existing work, our approach does \emph{not}
model the relation between $P_{g}$ and $P_{\text{data}}$, instead
we work directly with \emph{geometric} structure of the data. In particular,
we depart from the original idea of support vectors in \cite{benhur_etal_jmlr01_svc}
and formulate an optimization framework to ensure that the generated
value $G(\bz)$ will be contained within the data manifold learned
from training data. As in VAE or GAN, our approach is completely unsupervised,
hence to richly characterize this (nonlinear) data manifold, we borrow
the idea of constructing a \emph{minimal enclosing ball} $\mathcal{B}$
in the feature space whose theory guarantees the inverse mapping to
the data space produces nonlinear contours rich enough to describe
the data manifold \cite{benhur_etal_jmlr01_svc}. Our high-level intuition
is then to learn a generator $G(\bz)$ in such a way that its values,
when being mapped to the same feature space, are also enclosed in
the same ball $\mathcal{B}$, consequently $G\left(\bz\right)$ is
guaranteed to lie in the data manifolds. 

Directly using the primal form to construct $\mathcal{B}$ as in \cite{benhur_etal_jmlr01_svc}
is, however, not tractable for our purpose since we need the mapping
function $\Phi\left(\bx\right)$ from the input space to feature space
to be explicit. Furthermore, this function must facilitate efficient
parameter estimation procedure such as through backpropagation. We
overcome this obstacle by using recent Fourier random feature representation
proposed in \cite{rahimi_recht_nips07_random_feature} to approximate
$\Phi\left(\bx\right)$ with an explicit $\tilde{\Phi}$$\left(\bx\right)$
in a finite-dimensional space. To enable efficient learning via gradient
descent and backpropagation, we use the idea of reparameterization
in \cite{kingma_welling_iclr14_vae} to reformulate $\tilde{\Phi}$$\left(\bx\right)$.
Altogether, we arrive at an optimization framework which can be efficiently
solved via two stages: learn the enclosing ball $\mathcal{B}$ from
data via random feature reparameterization, followed by backpropagation
to train the generator $G$. We term our approach \emph{Geometric
Enclosing Networks} ($\model$) to reflect the fact that the spirit
of our approach is indeed corresponding to geometric intuition.

In addition, using the approximate mapping $\tilde{\Phi}$$\left(\bx\right)$
could however potentially result in a technical problem since the
theory of \cite{benhur_etal_jmlr01_svc} requires $\Phi(\bx$) to
be bijective. To this end, we provide theoretical analysis to guarantee
that the approximate mapping $\tilde{\Phi}\left(\bx\right)$ is a
bijection. In addition to the construction of $\model$, both the
random feature reparameterization and the bijection result for the
construction of minimal enclosing ball are also novel contributions,
to our knowledge. We conducted experiments to demonstrate the behaviors
of our proposed approach using synthetic and real-world datasets.
We demonstrate how our $\model$ can avoid mode collapsing problem
and result in better data generation quality via comparison with the
true data (when using synthesis data) and visual inspection on MNIST,
Frey Face, CIFAR-10, and CelebA datasets.

Compared with implicit density estimation approaches, epitomized by
GAN, our proposed $\model$ possesses some key advantages. First it
presents an orthogonal, fresh new idea to the problem of data generation
via geometric intuition as opposed to density estimation view. Second,
our optimization is simpler, much easier to control and it enjoys
the maturity of the optimization field at its hand. Third, it can
easily avoid the mode collapsing problem and efficiently learn manifold
in the data in a completely unsupervised manner. Lastly, it opens
up various promising future work on geometry-based approach to data
generation.

\vspace{-2mm}

\section{Related Background}

\vspace{-1mm}

Related closely to the technical development of our model is the theory
of support vectors and minimal enclosing ball \cite{Vapnik:1995,Cortes:1995:SN:218919.218929,benhur_etal_jmlr01_svc,tax2004support},
which we shall briefly describe in Section \ref{subsec:SVDD}. The
original primal form of \cite{Vapnik:1995,Cortes:1995:SN:218919.218929,benhur_etal_jmlr01_svc,tax2004support},
however, uses \emph{implicit} infinitely-dimensional mapping $\Phi\left(\cdot\right)$,
hence impeding the use of backpropagation and gradient descent training.
To overcome this obstacle, we briefly revise the Fourier random feature
representation of \cite{rahimi_recht_nips07_random_feature} in Section
\ref{subsec:Fourier-Random-Feature}. 

\vspace{-2mm}

\subsection{Minimal Enclosing Ball Optimization\label{subsec:SVDD}}

Given an unlabeled training set $\mathcal{D}=\left\{ \vectorize x_{1},\vectorize x_{2},...,\vectorize x_{N}\right\} $
where $\vectorize x_{i}\in\mathbb{R}^{d}$, \cite{tax1999support,tax2004support}
formulated an optimization to learn the data description via a minimal
enclosing ball $\mathcal{B}$ of the feature vectors $\left\{ \Phi\left(\vectorize x_{1}\right),\Phi\left(\vectorize x_{2}\right),\ldots,\Phi\left(\vectorize x_{N}\right)\right\} $
where $\Phi$ is a feature map from the input space to the feature
space:
\begin{align*}
 & \text{min}_{R,\vectorize c,\boldsymbol{\xi}}\left(\lambda R^{2}+\frac{1}{N}\sum_{i=1}^{N}\xi_{i}\right)\\
\text{s.t.}: & \norm{\Phi\left(\vectorize x_{i}\right)-\vectorize c}^{2}\leq R^{2}+\xi_{i},\,i=1,...,N;\,\xi_{i}\geq0,\,i=1,...,N
\end{align*}
where $R,\,\vectorize c$ are the radius and center of the minimal
enclosing ball respectively, $\boldsymbol{\xi}=\left[\xi_{i}\right]_{i=1}^{N}$
is the vector of slack variables, and $\lambda>0$ is the trade-off
parameter. Our work utilizes its primal form, which can be stated
as:{\small{}
\begin{equation}
\min_{R,\vectorize c}\left(\lambda R^{2}+\frac{1}{N}\sum_{i=1}^{N}\max\{0,\norm{\Phi\left(\vectorize x_{i}\right)-\vectorize c}^{2}-R^{2}\}\right)\label{eq:SVDD_primal_form}
\end{equation}
}{\small \par}

For our interest, an important result in \cite{tax1999support,tax2004support}
is that minimal enclosing ball in the feature space, when being mapped
back the input space, generates nonlinear contours that can be interpreted
as the clusters or data manifold of the training set. This principle
has been proven to be able to learn nested, or complicated clusters
and data manifolds in high dimensional space \cite{benhur_etal_jmlr01_svc}.

\vspace{-2mm}

\subsection{Fourier Random Feature Representation \label{subsec:Fourier-Random-Feature}\vspace{-2mm}
}

The mapping $\Phi\left(\vectorize x\right)$ above is implicitly defined
and the inner product $\left\langle \Phi\left(\vectorize x\right),\Phi\left(\vectorize x^{\prime}\right)\right\rangle $
is evaluated through a kernel $K\left(\vectorize x,\vectorize x^{\prime}\right)$.
To construct an explicit representation of $\Phi\left(\vectorize x\right)$,
the key idea is to approximate the symmetric and positive semi-definite
(p.s.d) kernel $K\left(\vectorize x,\vectorize x^{\prime}\right)=k\left(\bx-\vectorize x^{\prime}\right)$
with $K\left(\bzero,\bzero\right)=k\left(\bzero\right)=1$ using a
kernel induced by a random finite-dimensional feature map \cite{rahimi_recht_nips07_random_feature}.
The mathematical tool behind this approximation is the Bochner's theorem
\cite{bochner_2016_lectures}, which states that every shift-invariant,
p.s.d kernel $K\left(\vectorize x,\vectorize x^{\prime}\right)$ can
be represented as an inverse Fourier transform of a proper distribution
$p\left(\bomega\right)$ as below:
\begin{equation}
K\left(\vectorize x,\vectorize x^{\prime}\right)=k\left(\bu\right)=\int p\left(\bomega\right)e^{i\bomega^{\top}\bu}d\bomega\label{eq:dist_transform}
\end{equation}
 where $\bu=\vectorize x-\vectorize x^{\prime}$ and $i$ represents
the imaginary unit (i.e., $i^{2}=-1$). In addition, the corresponding
proper distribution $p\left(\boldsymbol{\omega}\right)$ can be recovered
through Fourier transform of kernel function as:
\begin{equation}
p\left(\bomega\right)=\left(\frac{1}{2\pi}\right)^{d}\int k\left(\bu\right)e^{-i\bu^{\top}\bomega}d\bu\label{eq:kernel_transform}
\end{equation}

Popular shift-invariant kernels include Gaussian, Laplacian and Cauchy.
For our work, we employ Gaussian kernel: $K(\bx,\bx')=k\left(\bu\right)=\exp\left[-\frac{1}{2}\bu^{\top}\Sigma\bu\right]$
parameterized by the covariance matrix $\Sigma\in\mathbb{R}^{d\times d}$.
With this choice, substituting into Eq.~(\ref{eq:kernel_transform})
yields a closed-form for the probability distribution $p\left(\bomega\right)$
which is $\mathcal{N}\left(\bzero,\Sigma\right)$.

This suggests a Monte-Carlo approximation to the kernel in Eq.~(\ref{eq:dist_transform}):
\begin{align}
K\left(\vectorize x,\vectorize x^{\prime}\right) & =\mathbb{E}_{\bomega\sim p\left(\bomega\right)}\left[\text{cos}\left(\bomega^{\top}\left(\vectorize x-\vectorize x^{\prime}\right)\right)\right]\approx\sideset{\frac{1}{D}}{_{i=1}^{D}}\sum\left[\cos\left(\bomega_{i}^{\top}\left(\vectorize x-\vectorize x^{\prime}\right)\right)\right]\label{eq:MCMC_approx}
\end{align}
where we have sampled $\boldsymbol{\omega}_{i}\iid\mathcal{N}\left(\boldsymbol{\omega}\gv\bzero,\Sigma\right)$
for $i\in\left\{ 1,2,...,D\right\} $.

Eq.~(\ref{eq:MCMC_approx}) sheds light on the construction of a
$2D$-dimensional random map $\tilde{\Phi}:\mathcal{X}\goto\mathbb{R}^{2D}$:
\begin{align}
\tilde{\Phi}\left(\vectorize x\right) & =\left[\frac{1}{\sqrt{D}}\cos\left(\bomega_{i}^{\top}\vectorize x\right),\frac{1}{\sqrt{D}}\sin\left(\bomega_{i}^{\top}\vectorize x\right)\right]_{i=1}^{D}\label{eq:original_RFF}
\end{align}
resulting in the approximate kernel $\tilde{K}\left(\vectorize x,\vectorize x^{\prime}\right)=\tilde{\Phi}\left(\vectorize x\right)^{\top}\tilde{\Phi}\left(\vectorize x^{\prime}\right)$
that can accurately and efficiently approximate the original kernel:
$\tilde{K}\left(\vectorize x,\vectorize x^{\prime}\right)\approx K\left(\vectorize x,\vectorize x^{\prime}\right)$
\cite{rahimi_recht_nips07_random_feature}. While we now have an explicit
mapping $\tilde{\Phi}\left(\vectorize x\right)$, it is not yet an
explicit function of $\Sigma$, hence does not facilitate learning
this kernel parameter via gradient descent. In Section \ref{subsec:Minimal-Enclosing-Ball},
we further develop reparameterized version of this random feature
representation such that $\tilde{\Phi}\left(\vectorize x\right)$
becomes an explicit function of the kernel parameter $\Sigma$, hence
can be learned from data. 
\vspace{-2mm}

\section{Geometric Enclosing Networks\label{sec:framework}\vspace{-2mm}
}

We present our proposed $\model$ in this section, starting with a
high-level geometric intuition. This is followed by detailed description
of the framework, its algorithm and implementation details. Finally,
theoretical analysis is presented.\vspace{-2mm}

\subsection{High-level Geometric Intuition\vspace{-1mm}
}

To strengthen the intuition described earlier, Figure \ref{fig:visualization_GMEB-1}
summarizes the key high-level intuition of our approach. Given training
data $\bx_{1},\ldots,\bx_{n}\in\mathcal{X}$ and assume there is an
explicit feature map $\tilde{\Phi}\left(\bx\right)$ from the input
space to feature space, our network first learns a \emph{minimal enclosing
ball} $\mathcal{B}$ of the training data samples in the random feature
space, i.e., the ball with minimal radius that encloses $\tilde{\Phi}\left(\bx_{1}\right),\ldots\tilde{\Phi}\left(\bx_{n}\right)$.
Next, we train a generator $G_{\psi}\left(\bz\right)$ such that its
generated samples $G_{\psi}\left(\bz_{i}\right)$, once being mapped
via $\tilde{\Phi}$ to produce $\tilde{\Phi}\left(G_{\psi}\left(\bz_{i}\right)\right)$
also fall into $\mathcal{B}$ where $\bz$ are drawn i.i.d from any
arbitrary noise space. So long as $\tilde{\Phi}$ is a bijective mapping
(as shown in Theorem \ref{theo:bijection}), the generated samples
$G\left(\vectorize z\right)=\tilde{\Phi}^{-1}\left(\tilde{\Phi}\left(G\left(\vectorize z\right)\right)\right)$
must locate in the contours characterizing the true data samples in
the input space. 

Geometrically, all points in the feature space including those mapped
from training data (green) and generated by the generator $G\left(\vectorize z\right)$
(yellow) are enclosed in the ball $\mathcal{B}$. Because $\tilde{\Phi}\left(\vectorize x\right)$
lies in the unit hypersphere in the random feature space for all $\vectorize x\in\mathcal{X}$,
the mapping of training data points and generated data points also
must lie on the surface of this unit hypersphere inside $\mathcal{B}$.
Those points at the section between $\mathcal{B}$ and the unit hypersphere
are called \emph{support vectors. }

Our optimization framework then consists of two sub-optimization problems:
learn the explicit mapping $\tilde{\Phi}$ and the minimal enclosing
ball $\mathcal{B}$ in the first step and then train a generator $G(\bz)$
in the second step which we detail in Sections \ref{subsec:Minimal-Enclosing-Ball}
and \ref{subsec:Learning-Generator} respectively.\vspace{-2mm}

\subsection{Learning $\tilde{\Phi}$ and $\mathcal{B}$ via Random Feature Reparameterization
\label{subsec:Minimal-Enclosing-Ball}\vspace{-1mm}
}

Recall that our first goal is to find a minimal enclosing ball $\mathcal{B}$
by solving the optimization problem in Eq.~(\ref{eq:SVDD_primal_form}).
However, in its primal form, the feature map $\Phi$ is unknown and
our approach seeks for an approximate \emph{explicit} mapping via
random feature representation as in Eq.~(\ref{eq:original_RFF}).
Although $\tilde{\Phi}\left(\vectorize x\right)$ has an explicit
form, it is represented indirectly through samples drawn from $p\left(\bomega\gv\Sigma\right)$,
hence it is still not possible to learn $\Sigma$ through gradient
descent yet. To do this, we need $\tilde{\Phi}\left(\vectorize x\right)$
to be a direct function of the kernel parameter $\Sigma$. Adopting the reparameterization method proposed in \cite{tu_etal_ijcai17_rrf}, we shift the source of randomness $\boldsymbol{\omega}\sim\mathcal{N}\left(\bomega\gv\bzero,\Sigma\right)$
to $\boldsymbol{\omega}=\bzero+L\be\,\text{where}\,\be\sim\mathcal{N}\left(\bzero,\boldsymbol{I}_{d}\right)\,\,\text{and}\,\,L=\Sigma^{1/2}$
so that the kernel in Eq.~(\ref{eq:dist_transform}) can now be re-written
as:\vspace{-8pt}
{\small{}
\begin{multline}
K\left(\vectorize x,\vectorize x^{\prime}\right)=k\left(\bu\right)=\int\mathcal{N}\left(\be\mid\bzero,\boldsymbol{I}_{d}\right)e^{i\left(L\be\right)^{\top}\bu}d\be\\
\,\,\,=\int\mathcal{N}\left(\be\mid\bzero,\boldsymbol{I}_{d}\right)\left[\cos\left(\transp{\be}L\boldsymbol{u}\right)+i\sin\left(\transp{\be}L\boldsymbol{u}\right)\right]d\be=\int\mathcal{N}\left(\be\mid\bzero,\boldsymbol{I}_{d}\right)\cos\left[\transp{\be}L\left(\vectorize x-\vectorize x^{\prime}\right)\right]d\be\label{eq:kernel_epsilon-1}
\end{multline}
}which can be again approximated as 
\begin{gather*}
K\left(\vectorize x,\vectorize x^{\prime}\right)\approx\tilde{K}\left(\vectorize x,\vectorize x^{\prime}\right)=\left\langle \tilde{\Phi}\left(\vectorize x\right),\tilde{\Phi}\left(\vectorize x^{\prime}\right)\right\rangle =\frac{1}{D}\sum_{i=1}^{D}\left[\cos\left(a_{i}\vectorize x\right)\cos\left(a_{i}\vectorize x^{\prime}\right)+\sin\left(a_{i}\vectorize x\right)\sin\left(a_{i}\vectorize x^{\prime}\right)\right]
\end{gather*}
where $\be_{i}\sim\mathcal{N}\left(\bzero,\boldsymbol{I}_{d}\right),\,a_{i}=\transp{\be_{i}}L,\,\forall i=1,\ldots,D$,
and the random feature map now has an explicit representation as well
as being a direct function of the kernel parameter $\Sigma$:
\begin{equation}
\tilde{\Phi}\left(\vectorize x\right)=D^{-1/2}\left[\cos\left(\transp{\be_{i}}L\vectorize x\right),\sin\left(\transp{\be_{i}}L\vectorize x\right)\right]_{i=1}^{D}\label{eq:new_repam_feature_map}
\end{equation}
Using this new reparameterized feature map for $\tilde{\Phi}$ in
Eq. (\ref{eq:new_repam_feature_map}), the optimization problem for
$\mathcal{B}$ in Eq.~(\ref{eq:SVDD_primal_form}) is now reformulated
as:
\begin{equation}
\text{min}_{\tilde{R},\tilde{\vectorize c},\Sigma}\mathcal{J}_{d}\left(\tilde{R},\tilde{\vectorize c},\Sigma\right)\label{eq:SVDD_primal_form-random-1}
\end{equation}
where we have defined the new objective function:
\[
\mathcal{J}_{d}\left(\tilde{R},\tilde{\vectorize c},\Sigma\right)=\lambda\tilde{R}^{2}+\frac{1}{N}\sum_{i=1}^{N}\max\left(0,\norm{\tilde{\Phi}\left(\vectorize x_{i}\right)-\vectorize{\tilde{c}}}^{2}-\tilde{R}^{2}\right)
\]
with $\tilde{\vectorize c}\in\mathbb{R}^{2D}$ is the center of $\mathcal{B}$
in the random feature space, and $\tilde{R}$ is its radius. We then
apply SGD to solve the optimization problem in Eq. (\ref{eq:SVDD_primal_form-random-1})
as summarized in Algorithm \ref{alg:GEN}.. It is worth noting that
we only learn a diagonal matrix $\Sigma$ for efficient computation.
\vspace{-2mm}

\subsection{Training Generator $\boldsymbol{G}\left(\cdot\right)$ via Backprop
\label{subsec:Learning-Generator}\vspace{-1mm}
}

We recruit a neural network $G$ parameterized by $\psi$ (i.e., $G_{\psi}$)
to model the generator. The noise sample $\vectorize z$ is i.i.d
sampled from the noise distribution $P_{\vectorize z}$ , which could
be any arbitrary noise distribution (e.g., uniform or Gaussian noise).
Given any sample $\vectorize z\sim P_{\vectorize z}$, we train the
generator such that $\tilde{\Phi}\left(G_{\psi}\left(\vectorize z\right)\right)$
falls into the ball $\mathcal{B}$ of $\left\{ \tilde{\Phi}\left(\vectorize x_{1}\right),...,\tilde{\Phi}\left(\vectorize x_{N}\right)\right\} $
in the random feature space obtained in the previous step. This reduces
to the following optimization:
\begin{equation}
\min_{\psi}\left(\mathcal{J}_{g}\left(\psi\right)=\ESS[\max(0,\norm{\tilde{\Phi}\left(G_{\psi}\left(\vectorize z\right)\right)-\tilde{\vectorize c}}^{2}-\tilde{R}^{2})]{P_{\bz}}\right)\label{eq:GMEB_OP}
\end{equation}
We train the generator $G\left(\cdot\right)$ in such a way that its
generated samples $G\left(\bz\right)$(s) spread over the surface
of the unit sphere lying in the ball $\mathcal{B}$. This ensures
the inverse mapping $G\left(\bz\right)=\tilde{\Phi}^{-1}\left(\tilde{\Phi}\left(G\left(\bz\right)\right)\right)$
to stretch out the data manifolds. However, minimizing the objective
function in Eq. (\ref{eq:GMEB_OP}) could lead to the fact that the
generator simply maps into a small region inside the ball $\mathcal{B}$,
which can induce a negligible cost. To overcome this issue, we augment
the objective function using the quantity $\norm{\mathbb{E}_{P_{\text{data}}}\left[\tilde{\Phi}\left(\vectorize x\right)\right]-\mathbb{E}_{P_{\vectorize z}}\left[\tilde{\Phi}\left(G\left(\vectorize z\right)\right)\right]}^{2}$
\cite{salimans2016improved}, which encourages the generator to produce
samples that spread over the ball $\mathcal{B}$. We now can employ
the backpropagation to learn $\psi$ via minimizing the augmented
objective function. \vspace{-2mm}

\subsection{Algorithm and Implementation\vspace{-1mm}
}

The key steps of $\model$ are presented in Algorithm \ref{alg:GEN}.
In the first phase (i.e., the first $L$ epochs), we learn the ball
$\mathcal{B}$ using training data mapped onto the random feature
space. It is worth noting that we update the variables $\tilde{R},\tilde{\vectorize c},\Sigma$
and $\psi$ using the mini-batches. However, for the sake of simplicity,
we present their single-point updates (cf. Lines 6, 7, 8, and 14 in
Algorithm \ref{alg:GEN}). In the second phase (i.e., the last $T-L$
epochs), we keep fixed the ball $\mathcal{B}$ and the kernel (i.e.,
keeping fixed $\tilde{R},\tilde{\vectorize c}$ and $\Sigma$) and
train the generator such that for every $\vectorize z\sim P_{\vectorize z}$,
the random feature image $\tilde{\Phi}\left(G_{\psi}\left(\vectorize z\right)\right)$
falls into the ball $\mathcal{B}$. \vspace{-2mm}

\subsection{Theoretical Analysis \vspace{-1mm}
}

In what follows, we present the theoretical analysis regarding to
the tightness of kernel approximation using Fourier random feature
with the reparameterization trick and the conditions under which the
original and random feature maps (i.e., $\Phi$ and $\tilde{\Phi})$
are bijections. \vspace{-1mm}
\begin{thm}
\label{thm:kernel_approx}With a probability at least $1-2^{7}\left(\frac{\text{diam}\left(\mathcal{X}\right)\norm{\Sigma^{1/2}}_{F}}{\theta}\right)^{2}\exp\left(\frac{-D\theta^{2}}{4\left(d+2\right)}\right)$
where we assume that $0<\theta\leq\text{diam}\left(\mathcal{X}\right)\norm{\Sigma^{1/2}}_{F}$,
$\text{diam}\left(\mathcal{X}\right)$ is the diameter of the compact
set $\mathcal{X}$, and $\norm{\Sigma^{1/2}}_{F}$ specifies the Frobenius
norm of the matrix $\Sigma^{1/2}$, we have the following inequality:
\[
\sup_{\vectorize x,\vectorize x^{\prime}\in\mathcal{X}}\left|\tilde{K}\left(\vectorize x,\vectorize x^{\prime}\right)-K\left(\vectorize x,\vectorize x^{\prime}\right)\right|<\theta
\]
\vspace{-2mm}
\begin{figure}[H]
\noindent \centering{}%
\begin{minipage}[t]{0.52\textwidth}%
\noindent \begin{center}
\includegraphics[width=1\columnwidth]{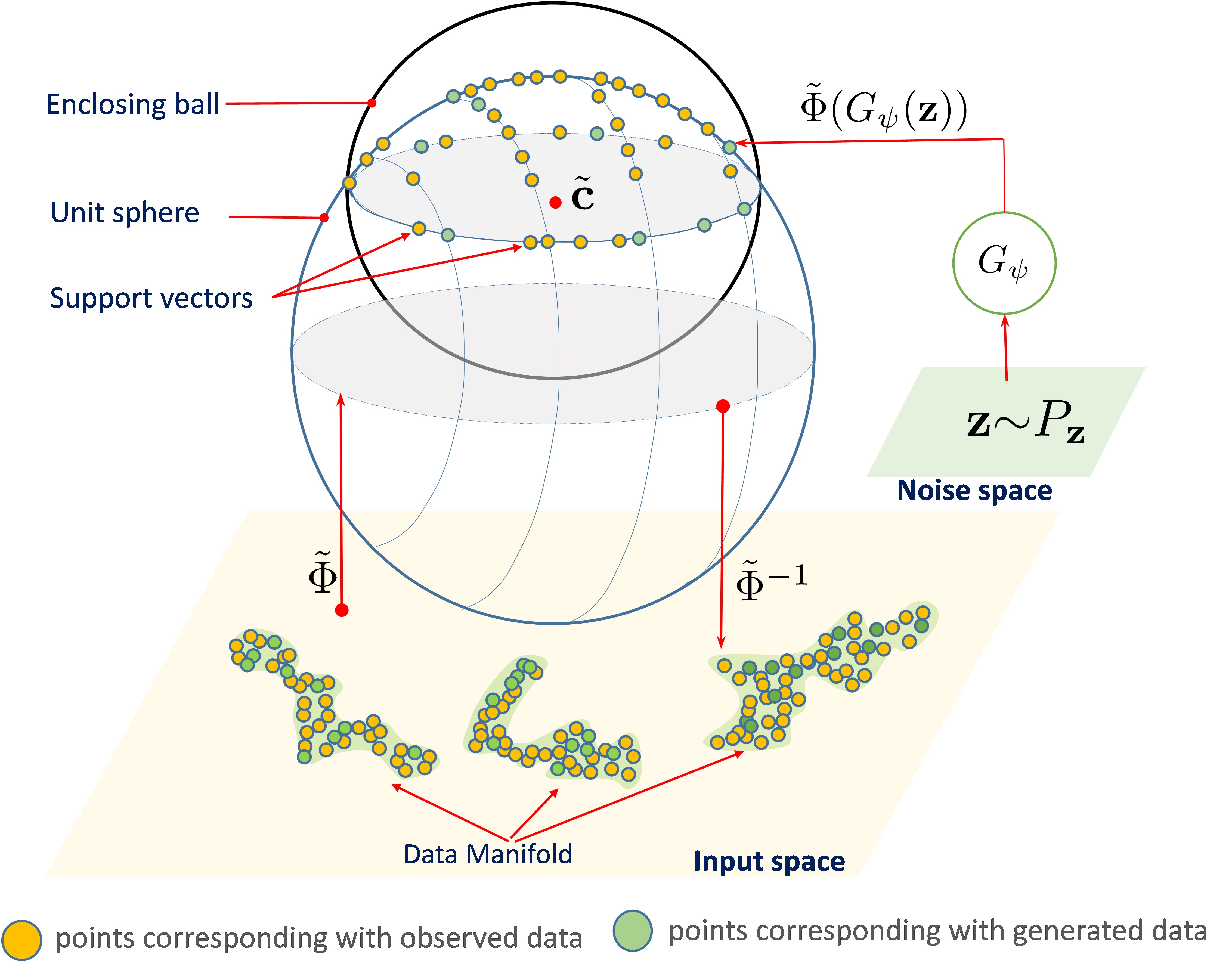}
\par\end{center}
\caption{Geometric structure of our proposed network ($\protect\model$). The
\emph{explicit }mapping $\tilde{\Phi}$ (and its inverse) is formulated
and learned via reparameterized random feature representation together
with the minimal enclosing ball $\mathcal{B}$. The generator $G\left(\protect\bz\right)$
is then trained via back propagation so that its image is contained
within $\mathcal{B}$. This ensures generated sample $G\left(\protect\vectorize z\right)$
belongs to the contours characterizing the data manifold in the input
space. Points at the intersection of $\mathcal{B}$ and the unit sphere
are called \emph{support vectors.} (best viewed in color).\label{fig:visualization_GMEB-1}}
\end{minipage}\hfill{}%
\begin{minipage}[t]{0.44\textwidth}%
\noindent \begin{center}
\vspace{4mm}
\par\end{center}
\captionof{algorithm}{Algorithm for GEN.}\label{alg:GEN}

\textbf{In}: $\lambda$, training data $\mathcal{D}=\left\{ \vectorize x_{i}\right\} _{i=1}^{N}$,
random feature dimension $D$

\textbf{Out}: $\tilde{R},\,\tilde{\vectorize c}$,~$\Sigma$,~$G_{\psi}$
\begin{algor}[1]
\item [{for}] $t=1$ \textbf{to} $T$
\item [{if}] $t\leq L$
\item [{{*}}] //\emph{ Learn the ball $\mathcal{B}$ }
\item [{for}] $n=1$ \textbf{to} $N$
\item [{{*}}] Sample \{$\vectorize x^{1},..,\vectorize x^{b}$\} from training
data $\mathcal{D}$
\item [{{*}}] $\tilde{R}=\tilde{R}-\eta\nabla_{\tilde{R}}\mathcal{J}_{d}\left(\cdot\right)$
\item [{{*}}] $\tilde{\vectorize c}=\tilde{\vectorize c}-\eta\nabla_{\tilde{\vectorize c}}\mathcal{J}_{d}\left(\cdot\right)$
\item [{{*}}] $\Sigma=\Sigma-\eta\nabla_{\Sigma}\mathcal{J}_{d}\left(\cdot\right)$
\item [{endfor}]~
\item [{else}]~
\item [{{*}}] // \emph{Fix $\tilde{R,}\tilde{\vectorize c},\Sigma$}
\item [{for}] $n=1$ \textbf{to} $N$
\item [{{*}}] Sample \{$\vectorize z^{1},..,\vectorize z^{b}$\} from the
noise distribution $P_{\vectorize z}$
\item [{{*}}] $\psi=\psi-\eta\nabla_{\psi}\mathcal{J}_{g}\left(\psi\right)$
\item [{endfor}]~
\item [{endif}]~
\item [{endfor}]~
\end{algor}
\end{minipage}\vspace{-2mm}
\end{figure}
\vspace{-4mm}
\end{thm}

From the representation of the random feature map $\tilde{\Phi}\left(\vectorize x\right)$,
it is obvious to verify that $\norm{\tilde{\Phi}\left(\vectorize x\right)}=\tilde{K}\left(\vectorize x,\vectorize x\right)^{1/2}=1$.
This is expressed through Lemma \ref{lem:norm_1-1} which helps us
construct the geometric view of our $\model$. 
\begin{lem}
\label{lem:norm_1-1}For every $\vectorize x\in\mathcal{X}\subset\mathbb{R}^{d}$,
the random feature map of $\vectorize x$ (i.e., $\tilde{\Phi}\left(\vectorize x\right)$)
lies in the unit hypersphere with the center origin and the radius
$1$.\vspace{-2mm}
\end{lem}

We are further able to prove that $\tilde{\Phi}:\,\mathcal{X}\goto\tilde{\Phi}\left(\mathcal{X}\right)$
is a bijective feature map if $\text{rank}\left\{ \be_{1},...,\be_{D}\right\} =d$
and $\norm{\Sigma^{1/2}}_{F}\text{diam}\left(\mathcal{X}\right)\max_{1\leq i\leq D}\norm{\be_{i}}<2\pi$
where $\text{diam}\left(\mathcal{X}\right)$ denotes the diameter
of the set $\mathcal{X}$ and $\norm{\Sigma^{1/2}}_{F}$ denotes the
Frobenius norm of the matrix $\Sigma^{1/2}$. This is stated in the
following theorem.\vspace{-1mm}
\begin{thm}
\label{theo:bijection}The following statements are guaranteed

i) The original feature map $\Phi:\,\mathcal{X}\goto\Phi\left(\mathcal{X}\right)$
is a bijective mapping.

ii) If $\Sigma$ is a non-singular matrix (i.e. positive definite
matrix), $\norm{\Sigma^{1/2}}_{F}\text{diam}\left(\mathcal{X}\right)\max_{1\leq i\leq D}\norm{\be_{i}}<2\pi$,
and $\text{rank}\left\{ \be_{1},...,\be_{D}\right\} =d$, then the
random map as in Eq. (\ref{eq:new_repam_feature_map})  $\tilde{\Phi}:\,\mathcal{X}\goto\tilde{\Phi}\left(\mathcal{X}\right)$
is a bijection feature map.\vspace{-2mm}
\end{thm}

Theorem \ref{theo:bijection} reveals that in order for the random
feature map $\tilde{\Phi}$ to be a bijection we can either set the
random feature dimension $D$ to a sufficiently large value such that
the random feature kernel $\tilde{K}\left(.,.\right)$ is sufficiently
approximate the original kernel $K\left(.,.\right)$ or scales the
data samples such that the inequality $\norm{\Sigma^{1/2}}_{F}\text{diam}\left(\mathcal{X}\right)\max_{1\leq i\leq D}\norm{\be_{i}}<2\pi$
is satisfied. The supplementary material further presents the proofs
for these claims.
\vspace{-2mm}

\section{Experimental Results\vspace{-2mm}
}

We conduct extensive experiments using both synthetic and real-world
datasets to demonstrate the properties of our proposed network; in
particular its ability to deal with multi-modal data distribution
(hence, avoiding mode collapsing problem), to model data manifold
and the quality of generated samples. Unless otherwise specified,
in all our experiments, when stochastic gradient descent was used,
the ADAM optimizer \cite{kingma2014adam} with learning rate empirically
turned by around 1e-3 and 1e-4 will be employed.\vspace{-2mm}

\subsection{Synthetic Data}

\vspace{-2mm}

\begin{figure}[H]
\begin{centering}
\includegraphics[width=1\textwidth]{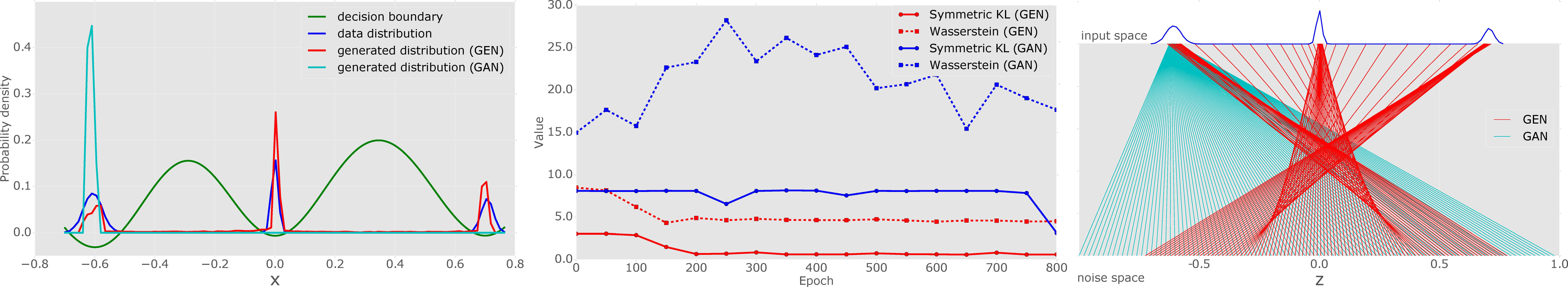}\vspace{-2mm}
\par\end{centering}
\caption{The comparison of $\protect\model$ and GAN on the 1D synthetic dataset.\label{fig:1D} }
\end{figure}

First, to see how well our $\model$ can deal with multiple modes
in the data, we generate 10,000 samples drawn from a mixture of univariate
Gaussians $0.45\times\mathcal{N}\left(-0.6,0.03\right)+0.25\times\mathcal{N}\left(0.7,0.02\right)+0.3\times\mathcal{N}\left(0,0.01\right)$
visualized as the blue curve on the left of Figure~\ref{fig:1D}.
Our baseline is GAN. The neural network specification for our generator
$G\left(\bz\right)$ includes 2 hidden layers, each with 30 softplus
units (and $D=100$ for the number of random features, cf. Eq.~(\ref{eq:new_repam_feature_map}))
and $\bz\sim\unipdf(-1,1)$. For GAN, we used a common setting \cite{goodfellow_etal_nips14_gan}
with one layer of 20 softplus hidden units for the generator, and
3 layers, each with 40 tanh hidden units for the discriminator.\vspace{-2.5mm}
 
\vspace{-5mm}

Figure~\ref{fig:1D} (left) shows pdfs (estimated from histograms
of $10,000$ data samples) generated by our $\model$ (red), GAN (cyan)
and the true distribution (blue). As it can clearly be seen, data
generated from our $\model$ distribute around \emph{all} three mixture
components, demonstrating its ability to deal with multi-modal data
in this case; whereas as expected, data generated from GAN concentrate
at a \emph{single} mode, reflecting the known mode collapsing problem
in GAN. In addition, we empirically observe that the pdf generated
by GAN is not stable as it tends to jump and fluctuate around the
$x$-axis during training, whereas ours is much more stable. \vspace{-6mm}
\begin{figure}[H]
\begin{centering}
\subfloat[Comparison of generated data from our $\protect\model$ and GAN in
2D case. True contours are blue. Ours can again capture multiple modes
(pink), whereas GAN failed to do so (green).\label{fig:2d_pdf}]{\begin{centering}
\includegraphics[width=0.45\textwidth,height=3cm]{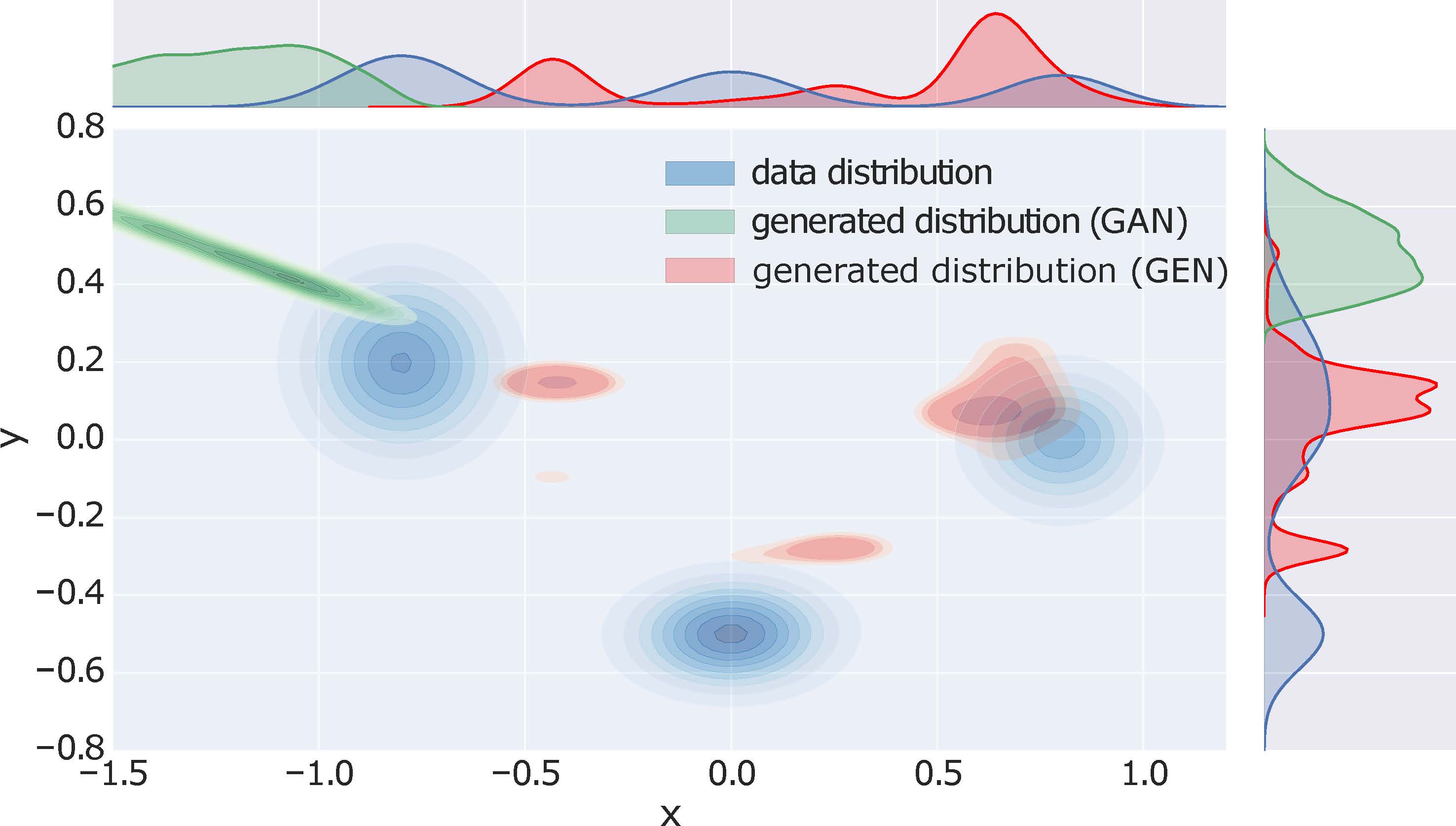}
\par\end{centering}
}~~~~~~~~~~~~~~~~~\subfloat[Data manifold captured by our model. Blue points are training data,
green curve is the boundary learned by our model via the discriminator
and red points are data generated by our model.\label{fig:2d-manifold}]{\begin{centering}
\includegraphics[width=0.45\textwidth,height=3cm]{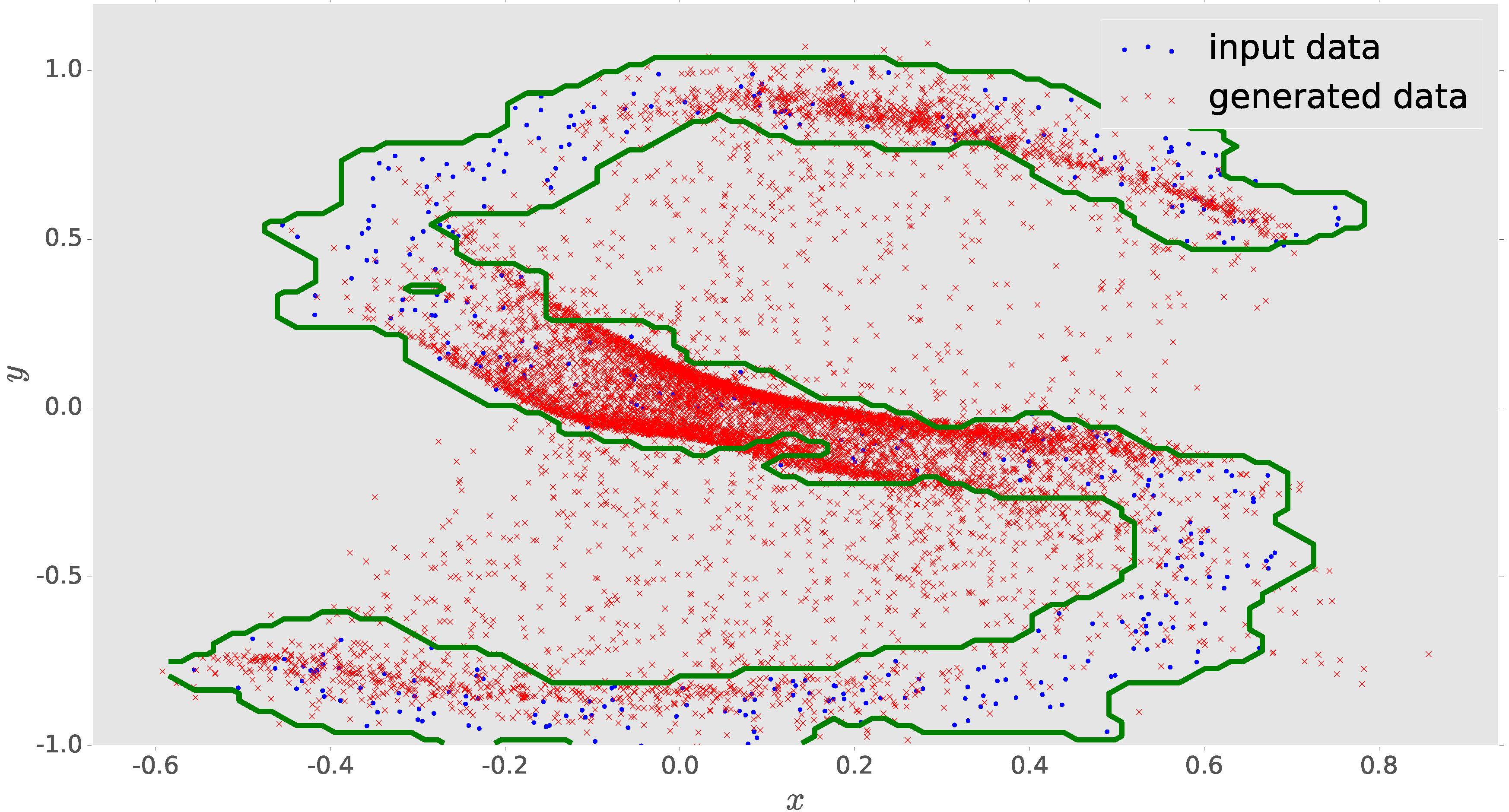}
\par\end{centering}
}\vspace{-2mm}
\par\end{centering}
\caption{Results on 2D synthetic data.}
\end{figure}
\vspace{-6mm}

Next, we further compare the quality of our generated data with that
of GAN quantitatively. Since we know the true distribution $P_{\text{data}}$
in this case, we employ two measures, namely symmetric KL and Wasserstein
distances. These measures compute the distance between the normalized
histograms generated from our GEN and GAN to the true $P_{\text{data}}$.
Figure~\ref{fig:1D} (middle) again clearly demonstrates the superiority
of our approach over GAN w.r.t. both distances; with Wasserstein metric,
the distance from ours to the true distribution almost reduces to
zero. This figure also demonstrates the stability of our $\model$
(red curves) during training as it is much less fluctuated compared
with GAN (blue curves).  Finally, Figure~\ref{fig:1D} (right) displays
the mapping from the noise space to the input data space learned by
$\model$ (red) and GAN (cyan). This again confirms that GAN is vulnerable
from the mode collapse whilst this is not the case for our method.

To further test our model beyond univariate case, we repeat a similar
setting on 2D case with a mixture of Gaussian whose means centered
at $\left[-0.8,0.2\right]$, $\left[0.8,0.0\right]$ and $\left[0.0,-0.5\right]$
respectively. Once again Figure~\ref{fig:2d_pdf} confirms that
our model can generate data at multiple locations (pink) whereas GAN
is stuck at one mode (green). To illustrate the idea of modeling data
manifold, we synthesize a data manifold having the S-shape as shown
in Figure~\ref{fig:2d-manifold}. The blue points represent the true
data samples. Using these samples, we train $\model$ and then generate
data shown as red points. It can be seen that the generated samples
spread out across the manifold as expected. In addition, the $\model$
discriminator can recognize the region of the S-shape (i.e., the region
covered by the green decision boundary).

\vspace{-3mm}

\subsection{Experiment with Real Datasets\vspace{-1mm}
}

\begin{figure}[H]
\begin{centering}
\includegraphics[width=1\textwidth]{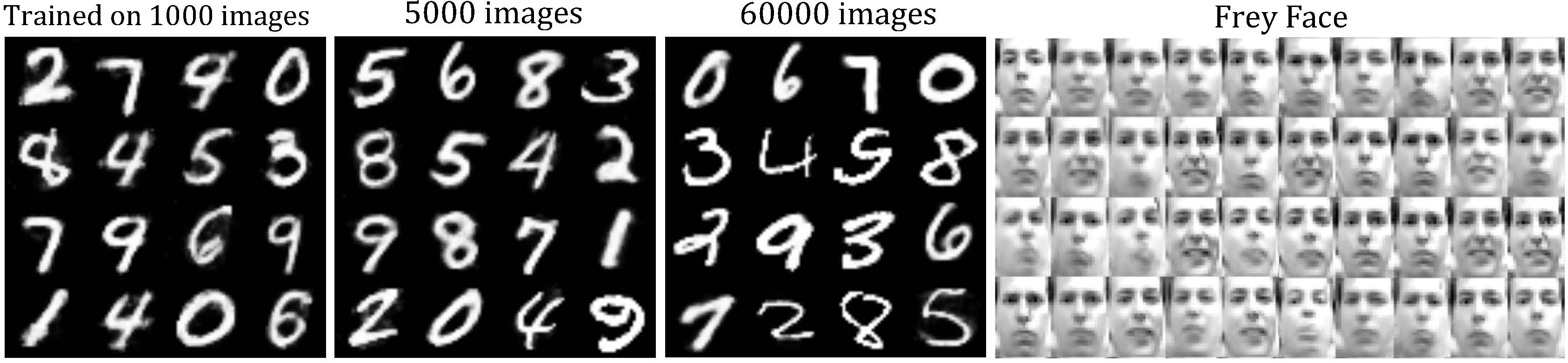}\vspace{-2mm}
\par\end{centering}
\caption{Data generated from our model when trained on MNIST dataset with $1,000$,
$5,000$, and $60,000$ training examples (left) and Frey Face dataset
(right).\label{fig:MNIST}}
\end{figure}

In this last experiment, we use four real-world image datasets: handwritten
digits (MNIST), object images (CIFAR-10) and two human face datasets
(Frey Face and CelebA). The popular MNIST dataset \cite{lecun_etal_IEEE98_gradient-based_learning}
contains $60,000$ images of digits from $0$ to $9$. We trained
our $\model$ with three subsets including $1,000$, $5,000$, and
$60,000$ images. The noise space for $\bz$ has 10 dimensions; our
generator $G\left(\bz\right)$ has the architecture of $1,000\rightarrow1,000\rightarrow1,000\rightarrow1,000$
(softplus units) and $784$ sigmoid output units; and $D=5,000$ random
features was used to construct $\tilde{\Phi}$. As can be visually
inspected from Figure~\ref{fig:MNIST} (left), the digits are generated
with good quality and improved with more training data. We note that
even with just $1,000$ training images, the quality is already sufficiently
good. This observation concurs with the synthetic experiments demonstrated
in Figure~\ref{fig:2d-manifold} where the geometric nature of our
$\model$ could usually generalize data manifold well with moderate
training data size. \vspace{-4mm}

\vspace{-8mm}

\begin{figure}[H]
\begin{centering}
\includegraphics[width=1\textwidth]{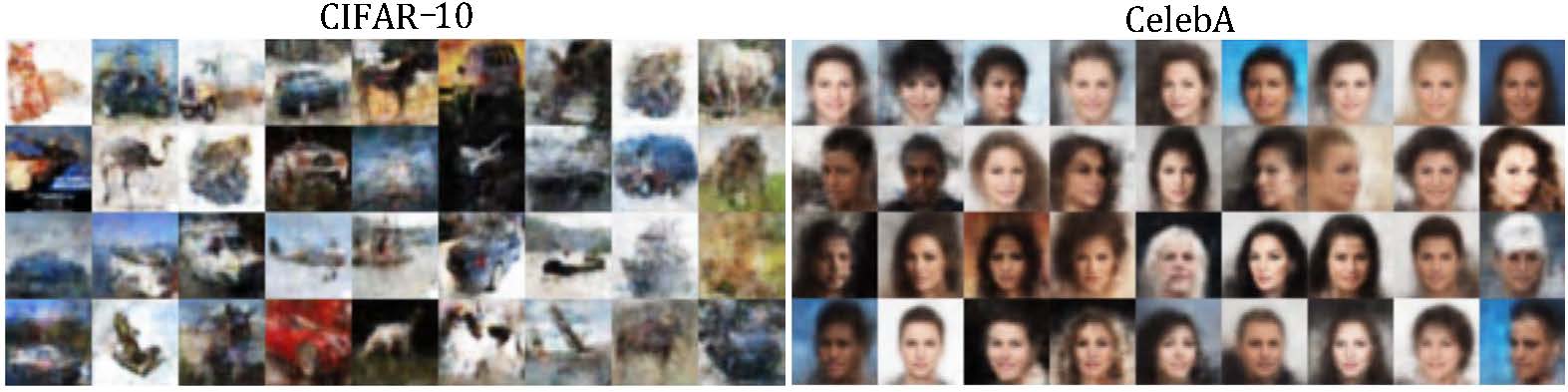}\vspace{-2mm}
\par\end{centering}
\caption{Data generated from our model when trained on CIFAR-10 dataset and
CelebA dataset.\label{fig:celev}}
\end{figure}
\vspace{-5mm}

The Frey Face dataset \cite{roweis_saul_Science00_nonlineardimensionality}
contains approximately 2000 images of Brendan's face, taken from sequential
frames of a small video. We used a smaller network for our generator
where $\bz$ has $5$ dimensions with a single layer of 200 softplus
hidden units and 560 sigmoid outputs. Figure~\ref{fig:MNIST} (right)
shows faces generated from our model where we can observe a good visual
quality.

Our experiments were further extended to generating color images of
real-life objects (CIFAR-10 \cite{krizhevsky_TR09learningmultiple})
and human faces (CelebA \cite{liu_etal_ICCV15_faceattributes}). After
resizing original images into the size of $32\times32\times3$, we
used a convolutional generator network with $512\rightarrow256\rightarrow128\rightarrow1,020$
(rectified linear units) and sigmoid output units and trained a leaky
rectified linear discriminator network with 3 layers $32\rightarrow64\rightarrow128$.
The random feature number and latent dimension are set to $5,000$
and $10$ respectively. We show the images generated from our GEN
network in Figure~\ref{fig:celev}. Although the generated images
are blurry and have low contrast, most of them contain meaningful
objects and faces in various shapes, colors and poses. These results
confirm the potential power of our proposed method as a generative
network.

\vspace{-3mm}

\section{Conclusion\vspace{-2mm}
}

This paper has presented a new approach, called Geometric Enclosing
Networks ($\model$), to construct data generator via geometric view.
Instead of examining the effectiveness of generator through the density
of generated data with the true (unknown) distribution, our approach
directly captures the nonlinear data manifolds in the input space
via the principle of minimal enclosing ball. As the result, our model
enjoys a nice geometric interpretation and possesses some key advantages,
namely simple and easy-to-control optimization formulation, avoidance
of mode collapse and efficiently learning data manifold representation.
We have established experiments to demonstrate the behaviors of our
proposed approach using synthetic and real-world datasets. The experimental
results show that our $\model$ can avoid mode collapsing problem
and result in better data generation quality via comparison with the
true data (when using synthesis data) and visual inspection on MNIST,
Frey Face, CIFAR-10, and CelebA datasets.

\clearpage{}

%\bibliographystyle{plain}
%\bibliography{GEN}

\section*{}
\end{document}